\documentclass[11pt]{article}

\usepackage[final]{acl}

\usepackage{times}
\usepackage{latexsym}

\usepackage[T1]{fontenc}
\usepackage[utf8]{inputenc}

\usepackage{microtype}

\usepackage{inconsolata}

\usepackage{graphicx}
\usepackage{float}
\usepackage{enumitem}
\usepackage{verbatim}
\usepackage{hyperref}
\usepackage{blindtext}
\usepackage{booktabs}
\usepackage{multirow}
\usepackage{makecell}
\usepackage{array}
\usepackage{subcaption}
\usepackage{amsmath}
\usepackage{paralist}

\title{Does Visual Rendering Bypass Tokenization? Investigating Script-Tokenizer Misalignment in Pixel-Based Language Models}
\author{
 \textbf{Lucky Susanto\textsuperscript{*1}},
 \textbf{Musa Izzanardi Wijanarko\textsuperscript{*1}},
 \textbf{Khumaisa Nur'aini\textsuperscript{1}},
 \textbf{Farid Adilazuarda\textsuperscript{4}},
\\
 \textbf{Alham Fikri Aji\textsuperscript{1,2}},
 \textbf{Derry Tanti Wijaya\textsuperscript{1,3}}
\\
\\
 \textsuperscript{*}Equal Contribution
\\
 \textsuperscript{1}Monash University Indonesia,
 \textsuperscript{2}MBZUAI,
 \textsuperscript{3}Boston University,
 \textsuperscript{4}University of Edinburgh
\\
 \small{
   \textbf{Correspondence:} \href{mailto:lucky.susanto@monash.edu}{lucky.susanto@monash.edu},
   \textbf{Correspondence 2:} \href{mailto:musa.wijanarko@monash.edu}{musa.wijanarko@monash.edu}
 }
}

\begin{document}
\maketitle
\begin{abstract}
While pixel-based language modeling aims to bypass the sub-word tokenization bottleneck by rendering text as images, recent multimodal variants such as DualGPT reintroduce text tokenizers to improve autoregressive performance. We investigate a fundamental question, \textbf{does visual rendering truly decouple a model from tokenization constraints?} Focusing on four Indonesian low-resource local languages that have their own non-Latin scripts (i.e., Javanese, Balinese, Sundanese, and Lampungnese), we evaluate the impact of script-tokenizer alignment within the DualGPT architecture. Our results show that, despite visual rendering, reintegrating a text tokenizer into the architecture reintroduces the same issue that pixel-based language modeling aims to resolve, which is the tokenizer misalignment problem. Despite having lower OOV and fertility rates, we show that the Llama 2 tokenizer performs significantly worse than a custom tokenizer, with improvements of up to 30.15 chrF++. Our findings serve as a warning for future multimodal variants, as text tokenizers remain a significant barrier to equitable models.
\end{abstract}

% This is a problem, as \textbf{these languages and their scripts} are tools for communication, becoming a vessel for history, identity, traditional knowledge, and social cohesion. Failure to solve this issue leads to digital marginalization and accelerates language attrition among many other things [ECIR 2024 Keynote].

\section{Introduction}

\begin{figure}[h]
  \includegraphics[width=\columnwidth]{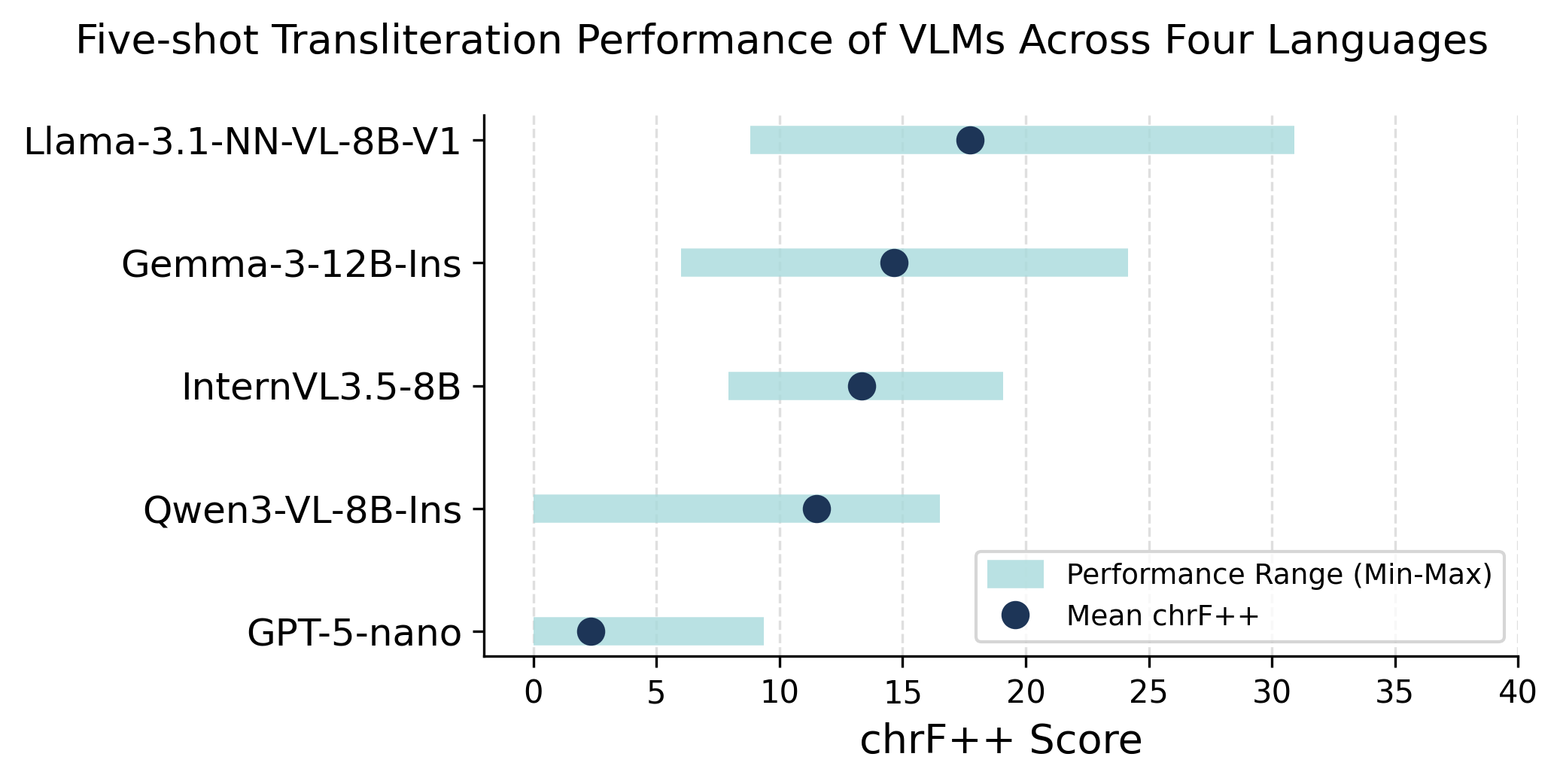}
  \caption{General VLMs image transliteration performance averaged across Javanese, Balinese, Sundanese, and Lampung on the NusaAksara evaluation dataset. Llama-3.1-NN-VL-8B-V1 is the Nemotron-Nano variant.}
  \vspace{-20pt}
  \label{fig:VLM-perf}
\end{figure}

While many Visual Language Models (VLMs) claim broad multilingual support (see Figure~\ref{fig:VLM-perf}; \citealp{bercovich2025llamanemotronefficientreasoningmodels,gemmateam2025gemma3technicalreport,wang2025internvl35advancingopensourcemultimodal,bai2025qwen3vltechnicalreport,openai2025gpt5systemcard}), their performance on low-resource languages with non-Latin scripts remains poor \citep{han2025mubenchassessmentmultilingualcapabilities}. A key challenge is script-tokenizer misalignment, where tokenizers trained predominantly on English data exhibit orthographic bias, causing excessive fragmentation of non-Latin scripts where single semantic units are split into multiple sub-word pieces.

Recent Pixel-based approaches introduced by  \citet{Rust2022LanguageMW} avoid tokenization by processing rendered text as images. PixelGPT \citep{chai-etal-2024-autoregressive} extends this with autoregressive variants, reintroducing text tokenizers in multimodal variants such as DualGPT. \textbf{If the visual modality truly decoupled the model from tokenization constraints, any tokenizer with high vocabulary coverage should yield similar results}. In contrast, our DualGPT results on four Indonesian low-resource local languages and scripts show that script-tokenizer misalignment remains a key bottleneck once tokenizers are reintroduced, even in pixel-based models.

\section{Background and Motivation}
\paragraph{Tokenization and Cross-lingual Disparities.} Current LLMs predominantly use BPE-based tokenization \citep{sennrich-etal-2016-neural, abagyan2025tokenizerruleallemergent}, including Llama 2 \citep{touvron2023llama2openfoundation}, which PixelGPT adopts \citep{chai-etal-2024-autoregressive}. However, BPE tokenizers trained on English-dominant corpora create inequities for underrepresented languages through excessive subword fragmentation \citep{foroutan2025parityawarebytepairencodingimproving, 10.5555/3666122.3667730}. Research on tokenization-free models \citep{clark-etal-2022-canine, xue-etal-2022-byt5} also offers an alternative approach for handling multilingual data. However, its application on generation tasks remains underexplored \citep{sun-etal-2023-multi}.

\paragraph{Pixel-based Language Modeling.} PIXEL \citep{Rust2022LanguageMW} addresses tokenization disparities by rendering text as images, theoretically enabling unlimited multilingual vocabulary. \citet{kesen-etal-2025-multilingual} demonstrate strong low-resource performance with this approach. However, rendering strategies affect patch diversity and learned representations \citep{lotz-etal-2023-text, tatariya-etal-2024-pixology}, with \citet{tatariya-etal-2024-pixology} noting that structured rendering may reintroduce tokenization-like constraints.

\paragraph{Multimodal Variants.} Recent work explores combining pixel-based and tokenization-based approaches. PixelGPT \citep{chai-etal-2024-autoregressive} extends PIXEL to autoregressive modeling, with its multimodal variant DualGPT integrating both visual rendering and text tokenization to leverage complementary strengths of each modality. While this hybrid approach demonstrates improved downstream performance, it raises an important question, \textbf{does reintroducing text tokenizers reintroduce the cross-lingual disparities that motivated pixel-based approaches?} We investigate this question by examining tokenizer alignment effects in DualGPT across low-resource Indonesian scripts and languages.

\section{Methodology}
\subsection{Datasets}
In this work, we analyze four Indonesian low-resource local scripts: Javanese (jav), Sundanese (sun), Balinese (ban), and Lampung (ljp), representing a range of resource levels and linguistic diversity.

\begin{table}[ht]
\centering
\small
\begin{tabular}{@{}lrrr@{}}
\toprule
\textbf{Language} & \textbf{\# Train} & \textbf{\# Eval}\\ \midrule
Javanese &  400,726 & 816 \\
Sundanese & 293,933 & 823\\
Balinese & 54,017 & 450 \\
Lampung & 945 & 84\\ \bottomrule
\end{tabular}
\caption{Dataset Statistics.}
\vspace{-10pt}
\label{tab:dataset_stat}
\end{table}

Training data comes from Wikidumps (July 2025) and digitalized Indonesian folklore\footnote{extracted using \texttt{pdfplumber}.} for Javanese, Sundanese, and Balinese, preprocessed as described in Section \ref{sec:tokenizer_renderer}. We use NusaAksara \citep{adilazuarda-etal-2025-nusaaksara} as the evaluation dataset and preprocess it in the same manner as the training data. Moreover, we split the NusaAksara's Lampung data for training and evaluation. Dataset statistics are presented in Table \ref{tab:dataset_stat}\footnote{Full dataset will be released after the double-blind review.}.

\subsection{PixelGPT}
We use DualGPT \citep{chai-etal-2024-autoregressive}\footnote{Weights taken from \texttt{ernie-research/DualGPT}}, pretraining the model on text, images, and paired text-image data, then finetuned for image-to-text transliteration. Each model went through a single training and finetuning run, with all reported metrics based on that execution.

\subsection{Tokenizer and Renderer}
\label{sec:tokenizer_renderer}
Both a tokenizer and a renderer are required as the DualGPT model relies on both text and image modalities during pre-training. The visualization for the data preprocessing pipeline can be seen in Appendix~\ref{sec:dataset_construction}.

\textbf{Tokenizer.} We train our models using two different tokenizers: the default tokenizer used in the PixelGPT architecture (Llama 2 tokenizer) and a custom-built tokenizer. Our tokenizer is inspired by a grapheme-based approach \citep{Basher2023BnGraphemizerAG}, rather than the default BPE-based method. First, we back-transliterate our training dataset from Latin script text into \textit{aksara} (the Indonesian local scripts) using community-built tools \footnote{Heuristic transliteration for \hyperlink{https://bennylin.github.io/transliterasijawa/}{Javanese}, \hyperlink{https://bennylin.github.io/transliterasi/bali.html}{Balinese}, and \hyperlink{https://sundadigi.com/konversi}{Sundanese}; and a custom font for \hyperlink{http://duniayuza.blogspot.com/2013/02/font-aksara-lampung-untuk-penulisan-di.html}{Lampung}.}. We then tokenize the aksara at the grapheme level and transliterate it back into Latin script text. Finally, we apply a word-level tokenization scheme to map the text into token IDs.

\textbf{Renderer.} With the findings of \citet{tatariya-etal-2024-pixology} and the nature of Javanese and Balinese, which do not mark word boundaries with whitespace, we adopt a continuous rendering strategy following \citet{lotz-etal-2023-text}. We use a custom Pillow-based renderer instead of the default Pygame renderer on our back-transliterated datasets, as the default renderer is unable to correctly render additional characters and diacritics. The renderer parameters can be seen in Appendix \ref{sec:renderer_params}.

% Kita tes 2 tokenizer: default tokenizer (Llama) dan custom-made tokenizer.
% @Izzan jelasin tokenizernya gimana buatnya. Jelasin kalau kita buat grapheme-based tokenizer dan berbasis latin text.

\subsection{Evaluation}
We use image transliteration as the evaluation task, which directly assesses script-tokenizer alignment by measuring surface-form preservation with minimal confounding variables. We report ChrF++ \citep{popovic-2017-chrf} in the main works, alongside BLEU \citep{papineni-etal-2002-bleu} and Word Error Rate in percentage (WER) in the Appendix (see Appendix~\ref{app:fullperf}).

\subsection{Model Setup}
Models are trained on 4×A40 GPUs with a maximum training time of 24 hours and early stopping (\texttt{patience = 5}) enabled. Training and finetuning hyperparameters are detailed in Appendix~\ref{app:hyperparams}.

% Each model is trained with 4xA40 GPU and early stopping is enabled with a patience of 5, using 1000 entries from each training data's language dataset used as validation set. Validation is performed every 1000 steps and the best model is used for finetuning. The finetuning setup is identical to the pretraining setup. Pretraining and Finetuning hyperparameters are available in [APPENDIX] and [APPENDIX] respectively.

\subsection{Tokenizer Statistics}
\label{sec:tokenizer_stat}
\begin{table}[ht]
\centering

\small
\begin{tabular}{@{}lrrr@{}}
\toprule
\textbf{Language} & \textbf{Llama 2} & \textbf{Custom} & \textbf{Inflation} \\ \midrule
Javanese$^\dagger$ & 45.97 & 65.11 & +41.6\% \\
Sundanese$^\ddagger$ & 39.41 & 61.42 & +55.8\% \\
Balinese$^\dagger$ & 130.04 & 130.86 & +0.6\% \\
Lampung$^\ddagger$ & 3.73 & 4.21 & +12.8\% \\ \bottomrule
\end{tabular}
\caption{Average text token length on train data. $^\dagger$Processed using the Javanese tokenizer. $^\ddagger$Processed using the Sundanese tokenizer.}
\label{tab:sequence_complexity}
\vspace{-0.6em}
\end{table}

\begin{figure}[b]
\centering
  \includegraphics[width=0.8\columnwidth]{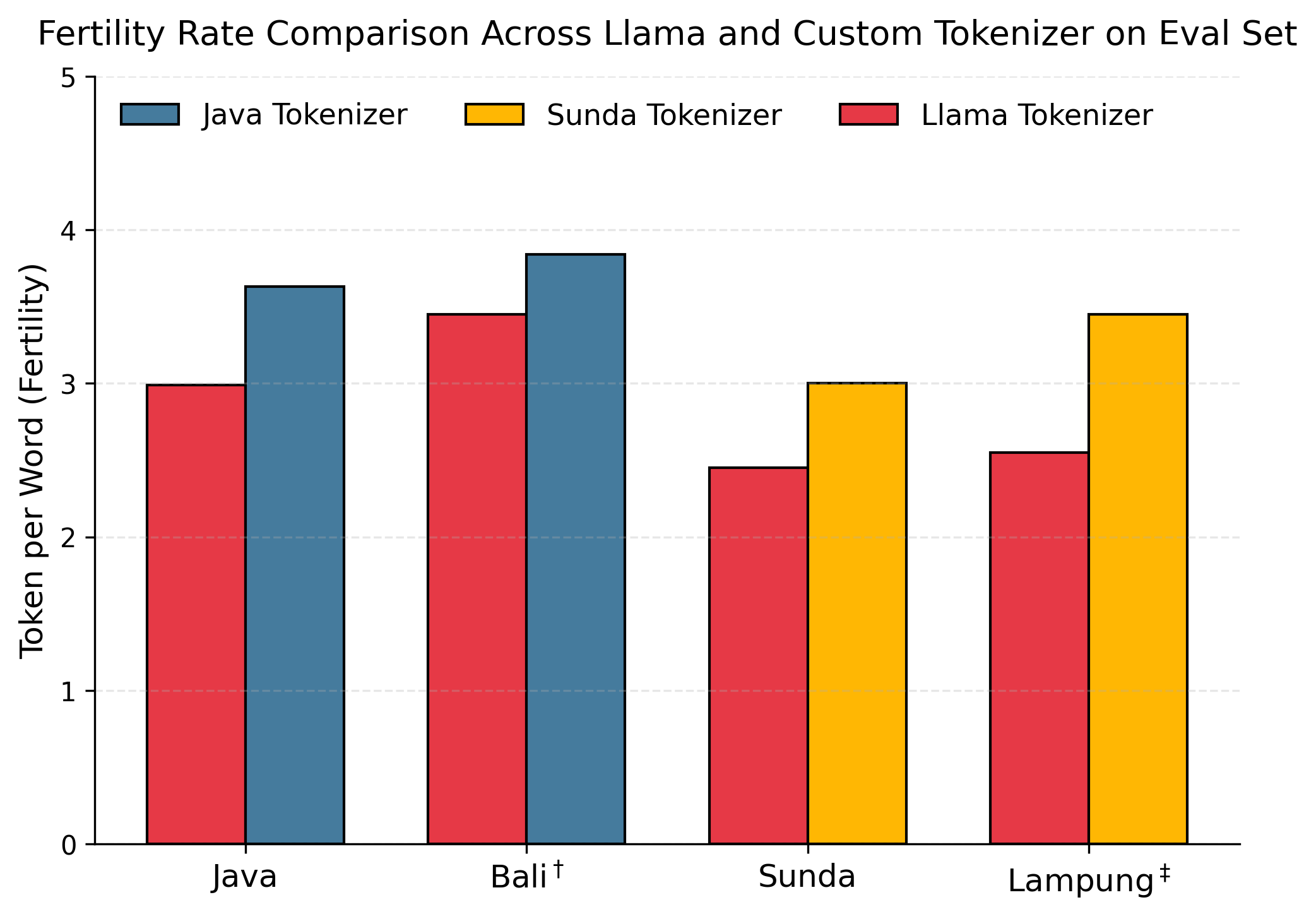}
  \caption{Fertility rate on the evaluation dataset per language. Bali$^\dagger$ and Lampung$^\ddagger$ uses Javanese's and Sundanese's tokenizer, respectively.}
  \label{fig:fertility_rate}
\end{figure}

Table \ref{tab:sequence_complexity} and Figure \ref{fig:fertility_rate} show that the Llama 2 tokenizer produces shorter sequences with lower fertility rates than custom tokenizers. The custom Javanese and Sundanese tokenizers achieve near-0\% Out-of-Vocabulary (OOV) rates for their target languages but exhibit 22\% and 10\% OOV rates on Balinese and Lampung, respectively. The Llama 2 tokenizer maintains 0\% OOV across all languages due to its broader vocabulary coverage.

By conventional metrics, sequence length, fertility rate, and OOV rate, the Llama 2 tokenizer appears superior. However, as we demonstrate in Section \ref{sec:results}, these efficiency metrics do not translate into real model performance on low-resource scripts.

% \begin{figure}[t]
%   \includegraphics[width=\columnwidth]{images/OOV_TokenRate_Eval.png}
%   \caption{Out of Vocabulary (OOV) token rate on the evaluation dataset per language. Bali* uses the Javanese Tokenizer, Lampung** uses the Sundanese Tokenizer.}
%   \label{fig:experiments}
% \end{figure}

% \begin{figure}[t]
%   \includegraphics[width=\columnwidth]{images/Tokenizer_FertilityRate_Eval.png}
%   \caption{Fertility rate on the evaluation dataset per language. Bali* uses the Javanese Tokenizer, Lampung** uses the Sundanese Tokenizer.}
%   \label{fig:fertility_rate}
% \end{figure}

\section{Results}
\label{sec:results}

\subsection{VLM Evaluation}
Five-shot in-context prompting reveals that current VLMs fail to transliterate rendered Indonesian scripts (Figure \ref{fig:VLM-perf}). While Llama-3.1-Nemotron-Nano-VL-8B-V1 achieves the highest chrF++ among tested models, its performance is negligible, with an average BLEU of 1.64 and a WER of 1,145.55 (many more errors than words in the reference). We report the full generation setup and results in Appendix~\ref{app:vlm-eval}.

\begin{figure}[tb]
  \includegraphics[width=\columnwidth]{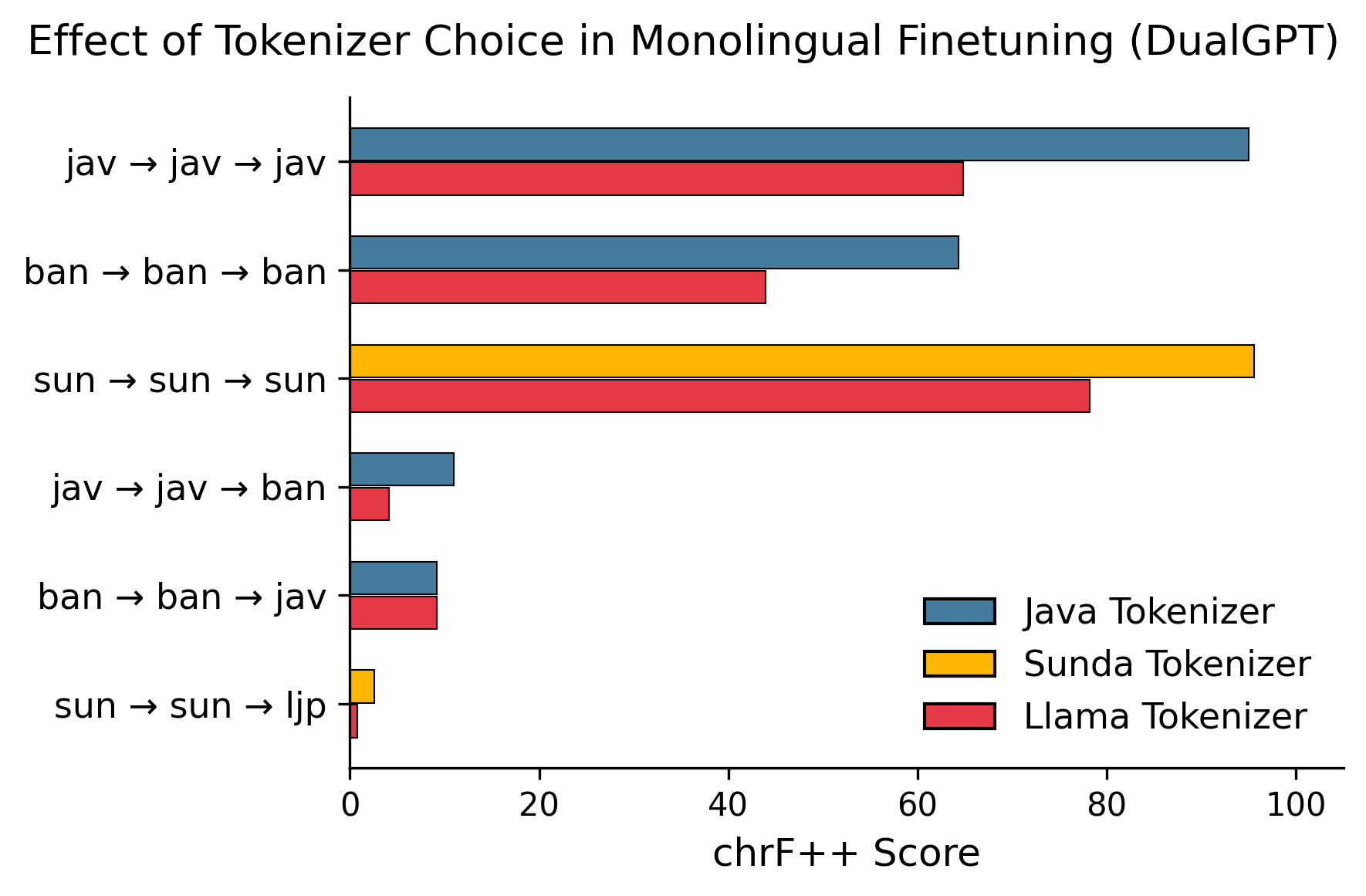}
  \caption{Impact of tokenizer choice on \textbf{Monolingual} DualGPT pretraining and finetuning: Pretraining $\rightarrow$ Finetuning $\rightarrow$ Evaluation.}
  \vspace{-10pt}
  \label{fig:tok-effect}
\end{figure}

\subsection{DualGPT Architecture}
Custom tokenizers significantly outperform Llama 2 in monolingual settings (Figure \ref{fig:tok-effect}), with chrF++ improvements of +30.15 for Javanese, +20.45 for Balinese, +17.40 for Sundanese, and +1.8 for Lampung (see Appendix~\ref{app:fullperf} for complete metrics). However, zero-shot cross-lingual transfer fails for both tokenizers (e.g., Java$\rightarrow$Bali only achieves 10.94 chrF++ with custom vs. 4.15 with Llama 2), indicating that improved in-language alignment does not enable cross-lingual generalization.

% Custom language-specific tokenizers significantly outperform the Llama baseline in the monolingual setup (see Figure \ref{fig:tok-effect}), yielding chrF++ gains of +30.15 (\%IMPROVE) for Javanese, +20.45 (\%IMPROVE) for Balinese, +17.40 (\%IMPROVE) for Sundanese, and +1.8 (\%IMPROVE) for Lampungnese. These improvements signify superior script-and-tokenizer alignment, signifying superior morphological preservation and lexical accuracy compared to the default Llama tokenizer. However, these gains are strictly in-domain, as the zero-shot cross-lingual transfer still fails regardless of tokenizer choice (e.g., Javanese$\rightarrow$Balinese reaching only 10.94 chrF++). This confirms that while high-granularity tokenization enhances intra-language fidelity, the resulting vocabulary misalignment acts as a barrier to inter-language generalization.

% COMMENTING FIGURE 4
% \begin{figure}[htb]
%   \includegraphics[width=\columnwidth]{images/tok_effect_multilingual.png}
%   \caption{Impact of tokenizer choice on \textbf{Multilingual} DualGPT pretraining and finetuning: Pretraining $\rightarrow$ Finetuning $\rightarrow$ Evaluation.}
%   \label{fig:tok-effect-multilingual}
% \end{figure}

\begin{table*}[ht]
\centering
\small
\begin{tabular}{@{}ccclrrr@{}}
\toprule
\textbf{Pretrain Data} & \textbf{Finetune Data} & \textbf{Eval Data} & \textbf{Tokenizer} & \textbf{ChrF++} $\uparrow$ & \textbf{WER} $\downarrow$ & \textbf{BLEU} $\uparrow$ \\ \midrule

% ===== JAVA =====
\multirow{4}{*}{Javanese}
 & \multirow{4}{*}{Javanese + Balinese}
 & \multirow{2}{*}{Javanese}
   & Llama  & 62.52 & 125.42 & 16.18 \\
 &  &  & Java & \textbf{94.27} & \textbf{21.79} & \textbf{78.05} \\
 \cmidrule(l){3-7}
 &  & \multirow{2}{*}{Balinese}
   & Llama  & 42.29 & \textbf{95.65} & 6.51 \\
 &  &  & Java & \textbf{64.11} & 135.62 & \textbf{7.92} \\ \midrule

% ===== BALI =====
\multirow{4}{*}{Balinese}
 & \multirow{4}{*}{Javanese + Balinese}
 & \multirow{2}{*}{Javanese}
   & Llama  & 62.48 & \textbf{74.84} & 23.88 \\
 &  &  & Java & \textbf{87.93} & 90.60 & \textbf{42.30} \\
 \cmidrule(l){3-7}
 &  & \multirow{2}{*}{Balinese}
   & Llama  & 42.64 & \textbf{105.35} & \textbf{5.98} \\
 &  &  & Java & \textbf{62.40} & 285.08 & 4.40 \\ \midrule

% ===== Javanese + Balinese =====
\multirow{4}{*}{\makecell[l]{Java + \\ Balinese}}
 & \multirow{4}{*}{Javanese + Balinese}
 & \multirow{2}{*}{Javanese}
   & Llama  & 64.06 & 127.51 & 17.33 \\
 &  &  & Java & \textbf{94.96} & \textbf{16.85} & \textbf{81.30} \\
 \cmidrule(l){3-7}
 &  & \multirow{2}{*}{Balinese}
   & Llama  & 44.89 & 159.35 & 5.11 \\
 &  &  & Java & \textbf{69.20} & \textbf{108.73} & \textbf{10.18} \\ \midrule

% ===== SUNDA =====
\multirow{4}{*}{Sundanese}
 & \multirow{4}{*}{Sunda + Lampung}
 & \multirow{2}{*}{Sundanese}
   & Llama  & 78.35 & 40.16 & 59.33 \\
 &  &  & Sunda & \textbf{96.02} & \textbf{5.33} & \textbf{92.45} \\
 \cmidrule(l){3-7}
 &  & \multirow{2}{*}{Lampung}
   & Llama  & 2.95 & 194.12 & 0.05 \\
 &  &  & Sunda & \textbf{9.67} & \textbf{159.85} & \textbf{0.13} \\ \bottomrule

\end{tabular}
\caption{Image transliteration performance comparing Llama and Custom tokenizers under a multilingual training setup. \textbf{Bolded} values indicate the best performance across tokenizers for each metric within each evaluation language. We use chrF++ as the main metrics for analysis (see Appendix~\ref{app:chrF}).}
\label{tab:multi_transliteration_result}
\vspace{-10pt}
\end{table*}
% (Figure \ref{fig:tok-effect-multilingual})
In multilingual training (Table~\ref{tab:multi_transliteration_result}), custom tokenizers maintain their advantage. Joint Java+Bali training with the Javanese tokenizer achieves 69.2 chrF++, % and 10.18 BLEU on Balinese
more than 6 times Java$\rightarrow$Bali %, nearly double the BLEU of 
monolingual training performance, suggesting that multilingual exposure improves cross-lingual transfer, though high WER (108.73\%) indicates continued word-level challenges.

Crucially, the Llama 2 tokenizer's 0\% OOV rate did not translate to superior performance. Despite having significantly higher OOV rates in cross-lingual settings (Section \ref{sec:tokenizer_stat}), our custom tokenizers maintained a substantial lead, suggesting that visual features do not compensate for poorly aligned text embeddings.

This script-tokenizer alignment problem is also evidenced by the Javanese and Balinese performance mismatch, although to a lesser extent than the Llama 2 tokenizer. Despite their shared Austronesian roots, Romanization conventions, and similar orthographic structures, the Javanese and Balinese models exhibit a significant chrF++ performance mismatch. Analyzing the vocabulary overlap, we find that a grapheme-based tokenizer trained on the Balinese data only has an overlap of 16.4\% with the used Javanese tokenizer (4,994 shared tokens out of 30,346 unique tokens). This indicates that the Balinese's performance mismatch stems from embedding space mismatch caused by the script-tokenizer misalignment rather than mere data scarcity.

These results demonstrate that tokenizer alignment critically affects performance, contradicting traditional efficiency metrics (Section \ref{sec:tokenizer_stat}): the Llama 2 tokenizer's superior fertility rate and zero OOV did not translate to better transliteration.

% \section{Analysis: Cross-lingual Tokenizer Mismatch}
% The Javanese and Balinese models in this work exhibit an extreme performance mismatch, despite sharing an Austronesian root, Romanization conventions, and similar orthographic structures. Our analysis reveals that even when using a grapheme-based tokenizer, meant to preserve morphological integrity, the vocabulary overlap between the two tokenizers is only 16.4\% (4,998 shared tokens out of a combined 30,436 unique tokens). As a result, there are tokens in Balinese that are not representable by the Javanese tokenizer, leading to a mismatch in the embedding space. Thus, rather than being caused by data scarcity, this ablation suggests that the Balinese model's poor performance is potentially affected by the tokenizer-and-script alignment mismatch.

% Note: bali_vocab_size: 15368 jawa_vocab_size: 20066 intersection: 4998

\section{Discussion and Future Works}
\paragraph{Tokenizer metrics are not indicative of tokenizer fit} While standard metrics (Section \ref{sec:tokenizer_stat}) suggest the Llama 2 tokenizer is superior for all four languages, each model performs significantly better with the custom tokenizer. We thus urge the creation of evaluation frameworks that better align tokenizer quality with model performance for both monolingual and multilingual cases \citep{chelombitko2024qtokcomprehensiveframeworkevaluating}.

\paragraph{Text tokenizer remains a bottleneck despite pixel-based modeling} One might expect that pixel-based architectures would not require a robust text tokenizer. However, we show that the text tokenizer remains a crucial bottleneck. While this work focuses on pixel-based models, current LLMs likely face similar issues with low-resource scripts, even when Romanized. We require further analysis on the impact of tokenizer choice on multilingual LLM performance \citep{limisiewicz-etal-2023-tokenization}.

\paragraph{Cross-lingual transfers are still limited by the tokenizer} Our analysis indicates that poor performance stems not only from data scarcity but from tokenizer mismatch. Despite Javanese and Balinese being linguistically highly related, cross-lingual transfer remains ineffective. This suggests that as long as the tokenizer issue persists, low-resource languages may benefit more from specialized models than from cross-lingual transfer. Advancements in architectures that balance specialization with generalization are vital for equitable performance.

\paragraph{Leveraging Rendered Representations} Despite our criticism, we believe pixel-based language modeling is a promising path toward more equitable technology. This work acts as a warning, reintroducing text tokenizers into such architectures may bring unintended consequences and must be done with proper justification and extreme care.

% Tokenizer masih bermasalah, dan kita menunjukkan ada high-chance kalau kita bikin smaller model with custom tokenizer, performanya bisa jauh lebih better daripada ICL dengan LLMs. Tapi on the other hand, custom tokenizer juga limit the generalization -izzan.

\section*{Limitations}
\paragraph{Scope of Architectural Analysis.} This work focuses specifically on the impact of tokenizer alignment within hybrid vision-text architectures (e.g., DualGPT). While a comparison with pure vision-based models (without text-heads) would provide additional context, our study is designed to isolate the behavior of the text-modality when reintroduced into pixel-based frameworks. Future work will expand this to a wider range of architectural configurations.

\paragraph{Resource Disparity.} The available training data for Javanese, Balinese, Sundanese, and Lampung varies significantly, which is reflected in the performance gap observed in the Lampung experiments. However, the consistent improvement across all languages when using aligned tokenizers suggests that our findings are robust to variations in data scale.

\paragraph{Task Specificity.} We utilize image transliteration as a primary probe because it directly measures surface-form preservation and script-alignment with minimal confounding variables. While this task effectively exposes the tokenizer bottleneck, further research is needed to determine how these alignment issues affect higher-level semantic tasks like NLI or abstractive summarization in pixel-based models.

% \paragraph{We do not have an image-only pre-training baseline yet.} This will be added in the camera-ready version.
% \paragraph{Dataset size is a free variable.} The dataset sizes for Javanese, Balinese, Sundanese, and Lampung are all different. This issue is very apparent when we report on Lampung performance. Despite this, we believe that the improvement exhibited across all languages is evident.
% \paragraph{Evaluation is limited to transliteration} Image transliteration works well as a task to probe for the tokenizer-and-script alignment problem. However, we have limited knowledge

% \section*{Ethical Consideration}
% This work relies solely on publicly available or properly licensed datasets and does not collect new personal or sensitive information. Text data are handled in line with common data protection principles, and the experiments use automatic metrics for system-level evaluation without involving human subjects, demographic profiling, or interventions.

% [1](https://digital-strategy.ec.europa.eu/en/library/ethics-guidelines-trustworthy-ai)

\section*{Acknowledgment}
Anonymized due to double blind.

\section*{Ethical Consideration}
Despite rendering native scripts as images, we still use the Romanized form to simplify the analysis in this work. This approach is not meant to supplant the use of native scripts with Latin texts, but is out of necessity, as many local Indonesian languages' scripts are not supported by Unicode. Further advancements in this line of research are pivotal to overcoming this digital infrastructure gap alongside improving language and script equity in the field of AI.

\bibliography{custom}

\appendix

\section{Hyperparameters}
\label{app:hyperparams}
Pretraining hyperparameters and finetuning hyperparameters are available in Table~\ref{tab:pretrain_hyperparam} and Table~\ref{tab:finetune_hyperparam}, respectively.

\begin{table}[h]
\centering
\small
\begin{tabular}{@{}ll r@{}}
\toprule
\textbf{Category} & \textbf{Hyperparameter} & \textbf{Value} \\ \midrule
\multirow{5}{*}{Architecture} & Model Backbone & DualGPT \\
 & Hidden / Interm. Size & 768 / 3072 \\
 & Layers / Attn Heads & 12 / 12 \\
 & Image Size & 16 $\times$ 16,384 \\
 & Patch Size & 16 \\ \midrule
\multirow{6}{*}{Optimization} & Optimizer & AdamW \\
 & Learning Rate & $5 \times 10^{-4}$ \\
 & LR Scheduler & Cosine \\
 & Warmup Steps & 1000 \\
 & Weight Decay & 0.1 \\
 & Numerical Precision & BF16 \\ \midrule
\multirow{7}{*}{Training} & Per-device Batch & 2 \\
 & Grad. Accumulation & 4 \\
 & Num. GPUs (A40) & 4 \\
 & Effective Batch Size & 32 \\
 & Max Training Time & 24 Hours \\
 & Early Stop Patience & 5 \\
 & Eval / Save Steps & 1000 \\ \bottomrule
\end{tabular}
\caption{Pre-training hyperparameters.}
\label{tab:pretrain_hyperparam}
\end{table}

\begin{table}[h]
\centering
\small
\begin{tabular}{@{}ll r@{}}
\toprule
\textbf{Category} & \textbf{Hyperparameter} & \textbf{Value} \\ \midrule
\multirow{3}{*}{Model Setup} & Pretrained Weight & DualGPT-pt \\
 & Dropout & 0.1 \\
 & Frozen Layers & \texttt{lm\_pixel\_head} \\ \midrule
\multirow{6}{*}{Optimization} & Learning Rate & $2 \times 10^{-5}$ \\
 & LR Scheduler & Cosine \\
 & Warmup Ratio & 0.03 \\
 & Weight Decay & 0.1 \\
 & Numerical Precision & BF16 \\ \midrule
\multirow{7}{*}{Training} & Per-device Batch & 2 \\
 & Grad. Accumulation & 4 \\
 & Num. GPUs (A40) & 4 \\
 & Effective Batch Size & 32 \\
 & Max Epochs & 10 \\
 & Early Stop Patience & 5 \\
 & Eval / Save Steps & 1000 \\ \bottomrule
\end{tabular}
\caption{Hyperparameters for image-to-text transliteration fine-tuning. We swap \textbf{Warmup steps} to \textbf{Warmup ratio} to account for Lampung finetuning data (945 entries). DualGPT-pt indicates the pretrained DualGPT model.}
\label{tab:finetune_hyperparam}
\end{table}

\section{Dataset Construction}
\label{sec:dataset_construction}

\begin{figure*}[h]
  \includegraphics[width=0.9\textwidth]{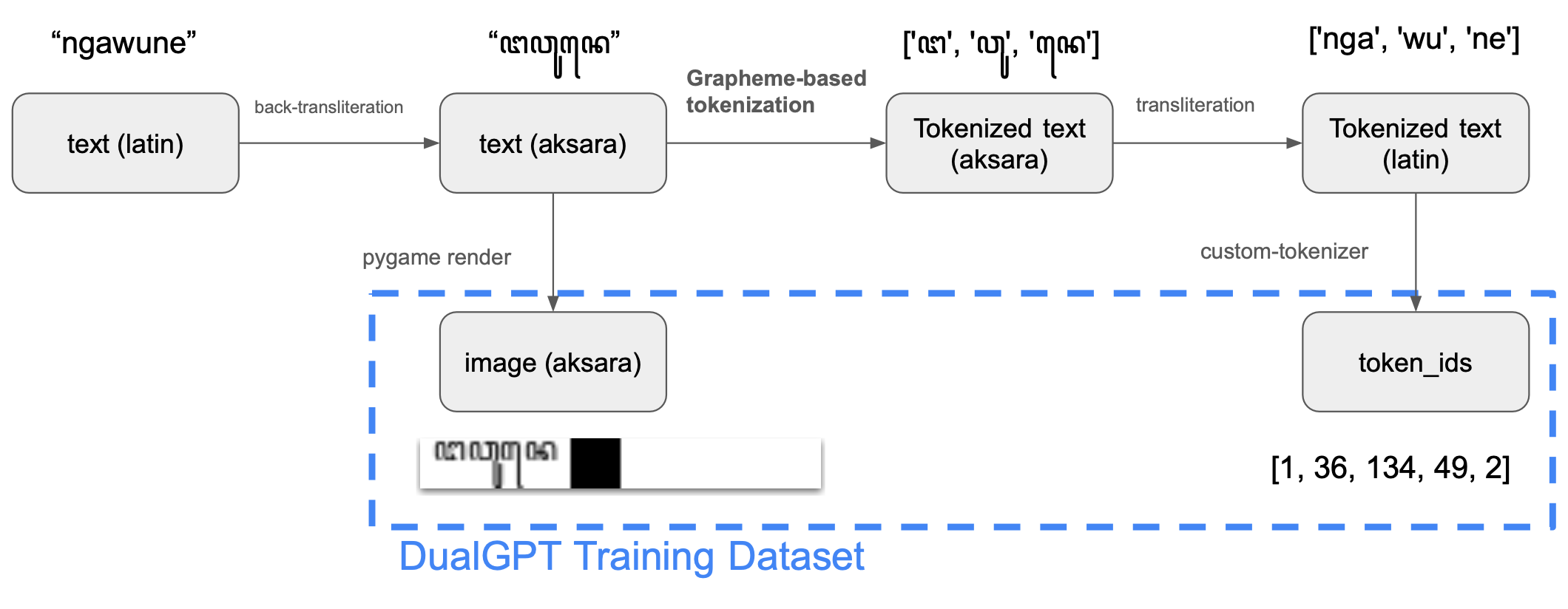}
  \caption{Dataset building process using custom tokenization rules.}
  \label{fig:builddataset}
\end{figure*}

Figure \ref{fig:builddataset} illustrates the dataset construction pipeline used for DualGPT training. Starting from romanized text, each word is back-transliterated into Indonesian scripts (Aksara), which is then tokenized using a grapheme-based scheme to preserve script-specific character structures. The Aksara text is rendered into an image using a custom renderer, forming the visual modality, while the tokenized Aksara is transliterated back into Latin and converted into token IDs using a custom tokenizer. The resulting paired image–token representation constitutes the final training dataset.

\section{Renderer Parameters}
\label{sec:renderer_params}
We follow the original PixelGPT renderer parameters, modifying only the font and font size to prevent text from overflowing beyond the image boundaries as shown in Table \ref{tab:text_render_config1}. Lampung language used  \hyperlink{http://aksaralampung.blogspot.com/2013/08/had-lampung-yuza-rounded.html}{custom font} since it is not inherently suppported by Unicode.

\begin{table}[h]
\small
\centering
\begin{tabular}{l l p{2cm}}
\hline
\textbf{Parameter} & \textbf{jav/sun/ban} & \textbf{ljp}\\
\hline
Background Color & White & White\\
DPI & 120 & 120\\
Font Color & Black & Black\\
Font Type & Noto Sans & Had Lampung Yuzu Rounded\\
Font Size & 7 & 10\\
Max sequence length & 1024 & 1024 \\
Padding size & 3 & 3\\
Pixels per patch & $16 \times 16$ & $16 \times 16$ \\
\hline
\end{tabular}
\caption{Configuration of text rendering for Javanese, Sundanese, Balinese, and Lampung language.}
\label{tab:text_render_config1}
\end{table}

\section{VLM Evaluation Setups and Results}
\label{app:vlm-eval}

We use open small-sized to medium-sized VLMs, including \texttt{Gemma-3-12B-it} \citep{gemmateam2025gemma3technicalreport}, \texttt{Qwen3-VL-8B-Instruct} \citep{bai2025qwen3vltechnicalreport}, \texttt{InternVL3.5-8B} \citep{wang2025internvl35advancingopensourcemultimodal}, and \texttt{Llama-3.1-Nemotron-Nano-8B-V1} \citep{bercovich2025llamanemotronefficientreasoningmodels}, as well as a closed-source model, GPT-5-nano \citep{openai2025gpt5systemcard}. Few-shot examples are randomly selected from pretrain data to avoid leakage. For the open models, we deploy them using vLLM \citep{kwon2023efficient}. For the generation hyperparameters, we constrain \texttt{temperature = 0.4} and \texttt{max\_output\_tokens = 2048}. Full VLMs evaluation performance is available at Table~\ref{tab:vlm-full-results}.

\section{Full Monolingual Model Performance}
\label{app:fullperf}
Full model monolingual performance is available in Table~\ref{tab:mono_transliteration_results}.

\section{chrF++ as the Main Metric}
\label{app:chrF}
Due to the nature of languages we are working with (i.e., low resource, native scripts), finding a suitable dataset for the evaluation remains a significant challenge. NusaAksara exist as one of the only benchmark available to support our transliteration task. However, NusaAksara has limitations. Shown in Table~\ref{tab:translit_samples}, NusaAksara's Balinese entry consist mainly of short word entries, disadvantaging metrics such as BLEU and WER.

\begin{table}[h]
\centering
\small
\begin{tabular}{@{}c ll@{}}
\toprule
\textbf{ID} & \textbf{Reference} & \textbf{Prediction} \\ \midrule
7 & \texttt{prathiwitala} & \texttt{prathiwitala} \\
9 & \texttt{mandaaakranta} & \texttt{man\textbf{t}aaakran\textbf{b}a} \\
11 & \texttt{134 ,} & \texttt{13 ,} \\
13 & \texttt{dumala} & \texttt{\textbf{mera}lumala} \\ \bottomrule
\end{tabular}
\caption{Transliteration sample for Balinese language. ID refers to NusaAksara's dataset entry ID.}
\label{tab:translit_samples}
\end{table}

\begin{table*}[t]
\centering
\resizebox{\textwidth}{!}{%
\begin{tabular}{@{}l rrr rrr rrr rrr@{}}
\toprule
\multirow{2}{*}{\textbf{Model}} & \multicolumn{3}{c}{\textbf{Balinese}} & \multicolumn{3}{c}{\textbf{Javanese}} & \multicolumn{3}{c}{\textbf{Lampung}} & \multicolumn{3}{c}{\textbf{Sundanese}} \\ \cmidrule(lr){2-4} \cmidrule(lr){5-7} \cmidrule(lr){8-10} \cmidrule(lr){11-13}
 & BLEU $\uparrow$ & chrF++ $\uparrow$ & WER $\downarrow$ & BLEU $\uparrow$ & chrF++ $\uparrow$ & WER $\downarrow$ & BLEU $\uparrow$ & chrF++ $\uparrow$ & WER $\downarrow$ & BLEU $\uparrow$ & chrF++ $\uparrow$ & WER $\downarrow$ \\ \midrule
InternVL3.5-8B & 0.00 & 9.18 & 564.94 & 0.00 & 17.10 & 1947.98 & 0.00 & 7.92 & 625.11 & 6.57 & 19.08 & 1444.19 \\
Llama-3.1-Nemotron-Nano-VL-8B-V1 & 4.99 & \textbf{19.44} & 794.22 & 0.52 & 8.82 & 907.54 & \textbf{1.09} & \textbf{11.83} & 2086.88 & 7.81 & \textbf{30.91} & 1200.58 \\
Qwen3-VL-8B & 1.62 & 14.21 & 238.35 & 2.91 & 16.50 & 316.89 & 0.00 & 0.00 & 1447.06 & \textbf{8.12} & 15.34 & 710.77 \\
Gemma-3-12B-it & \textbf{6.57} & 15.26 & 295.63 & \textbf{3.83} & \textbf{24.15} & 625.47 & 0.00 & 6.01 & 225.57 & 5.52 & 13.20 & 519.27 \\
GPT-5-Nano & 0.00 & 0.00 & \textbf{205.42} & 0.00 & 0.00 & \textbf{161.94} & 0.00 & 9.36 & \textbf{101.81} & 0.00 & 0.00 & \textbf{123.71} \\ \bottomrule
\end{tabular}
}
\caption{VLM Zero-shot image transliteration performance across four Indonesian languages. Higher is better for BLEU and chrF++; lower is better for WER. Note that extremely low WER in models with zero BLEU (e.g., GPT-5-Nano) typically indicates empty or extremely short output rather than accuracy.}
\label{tab:vlm-full-results}
\end{table*}

\begin{table*}[h]
\centering
\begin{tabular}{@{}cllrrr@{}}
\toprule
\textbf{Pretrain \& Finetune} & \textbf{Eval} & \textbf{Tokenizer} & \textbf{ChrF++} $\uparrow$ & \textbf{WER} $\downarrow$ & \textbf{BLEU} $\uparrow$ \\ \midrule

\multirow{4}{*}{Javanese} 
 & \multirow{2}{*}{Javanese} 
   & Llama  & 64.83 & 79.01 & 24.99 \\
 &  & Java & \textbf{94.98} & \textbf{19.95} & \textbf{80.28} \\ 
 \cmidrule(l){2-6}
 & \multirow{2}{*}{Balinese} 
   & Llama  & 4.10 & \textbf{506.99} & 0.01 \\
 &  & Java & \textbf{10.94} & 550.00 & 0.01 \\ \midrule

\multirow{4}{*}{Balinese} 
 & \multirow{2}{*}{Javanese} 
   & Llama  & \textbf{9.20} & 250.00 & 0.03 \\
 &  & Java & 9.16 & \textbf{102.30} & 0.03 \\ 
 \cmidrule(l){2-6}
 & \multirow{2}{*}{Balinese} 
   & Llama  & 43.91 & \textbf{86.16} & \textbf{7.53} \\
 &  & Java & \textbf{64.36} & 131.91 & 7.00 \\ \midrule

\multirow{4}{*}{Sundanese} 
 & \multirow{2}{*}{Sundanese} 
   & Llama  & 78.19 & 40.54 & 58.93 \\
 &  & Sunda & \textbf{95.59} & \textbf{7.87} & \textbf{90.55} \\ 
 \cmidrule(l){2-6}
 & \multirow{2}{*}{Lampung} 
   & Llama  & 0.77 & \textbf{325.79} & 0.00 \\
 &  & Sunda & \textbf{2.57} & 585.64 & 0.00 \\ \bottomrule

\end{tabular}
\caption{Image transliteration performance comparing Llama and Custom tokenizers under a monolingual tranining setup. Bold values indicate which tokenizer performs better.}
\label{tab:mono_transliteration_results}
\end{table*}

\end{document}